\documentclass[10pt,twocolumn,letterpaper]{article}
\usepackage[pagenumbers]{templates/EV24/cvpr}
\usepackage[table, dvipsnames]{xcolor}

\usepackage{graphicx}
\usepackage{amsmath}
\usepackage{amssymb}
\usepackage{booktabs}
\usepackage{enumitem}
\usepackage{xfrac}
\usepackage{adjustbox}
\usepackage{svg}
\usepackage{style}
\usepackage{pifont}
\usepackage{multirow}
\usepackage{url}
\usepackage{wrapfig}
\usepackage{mwe}
\usepackage{transparent}
\usepackage{cuted}
\usepackage{float}
\usepackage{microtype}
\definecolor{cvprblue}{rgb}{0.21,0.49,0.74}
\usepackage[pagebackref,breaklinks,colorlinks,citecolor=cvprblue]{hyperref}

\def\confName{ISPRS}

\usepackage[capitalize]{cleveref}
\crefname{section}{Sec.}{Secs.}
\Crefname{section}{Section}{Sections}
\Crefname{table}{Table}{Tables}
\crefname{table}{Tab.}{Tabs.}

\title{Satellite Image Time Series Semantic Change Detection:\\ Novel Architecture and Analysis of Domain Shift}

\author{Elliot Vincent\textsuperscript{1, 2}\\
\and
Jean Ponce\textsuperscript{3, 4}\\
\and
Mathieu Aubry\textsuperscript{1}\\
\and
{{\textsuperscript{1}LIGM, Ecole des Ponts, Univ Gustave Eiffel, CNRS, France}}\\
{{\textsuperscript{2}Inria Paris}}\\
{{\textsuperscript{3}Department of Computer Science, Ecole normale supérieure (ENS-PSL, CNRS, Inria)}}\\
{{\textsuperscript{4}Courant Institute of Mathematical Sciences and Center for Data Science, New York University}}
}

\begin{document}
\maketitle

\begin{abstract}
   Satellite imagery plays a crucial role in monitoring changes happening on Earth's surface and aiding in climate analysis, ecosystem assessment, and disaster response. In this paper, we tackle semantic change detection with satellite image time series (SITS-SCD) which encompasses both change detection and semantic segmentation tasks. We propose a new architecture that improves over the state of the art, scales better with the number of parameters, and leverages long-term temporal information. However, for practical use cases, models need to adapt to spatial and temporal shifts, which remains a challenge. We investigate the impact of temporal and spatial shifts separately on global, multi-year SITS datasets using DynamicEarthNet~\cite{Toker_2022_CVPR} and MUDS~\cite{van2021multi}. We show that the spatial domain shift represents the most complex setting and that the impact of temporal shift on performance is more pronounced on change detection than on semantic segmentation, highlighting that it is a specific issue deserving further attention. Our complete code is available at \url{https://github.com/ElliotVincent/SitsSCD}.

\end{abstract}

\section{Introduction}

The surface of the Earth is subject to constant changes, caused by human activity, natural disasters, and many other phenomena. As Earth observation from space has become widely accessible, it is acknowledged as "the most crucial input"~\cite{wmo2024} and "the best measure available"~\cite{hascic2018} for climate, ecosystem, and biodiversity monitoring. For example, it has proven useful in assessing flood risks in Italy~\cite{Moumtzidou_2020_eopen}, providing food security-related insights in South Korea~\cite{Moumtzidou_2020_eopen} and responding to wildfires in Australia~\cite{bushfire2020} or cyclones in New Zealand~\cite{linz2023}. The goal of this paper is to better improve and better understand the challenges in satellite image time series semantic change detection (SITS-SCD), \textit{i.e.} the detection of change in land use and land cover over time. We introduce an architecture that significantly boosts SITS-SCD results in the absence of any particular domain shift. However, practical monitoring necessitates online, real-time analysis, requiring models to accommodate the temporal shift between data seen during training and at inference. Additionally, due to the scarcity of annotated data~\cite{ball2017comprehensive, reichstein2019deep, bai2023deep}, many models in practical applications are applied to images gathered from places far away from where the training data was observed. For these reasons, we conduct a comprehensive analysis of the impact of spatial and temporal domain shifts, showing their critical significance in this context.

Many works are dedicated to addressing the spatio-temporal shift through domain adaptation. While much of this work concentrates on spatial domain adaptation for single satellite images~\cite{deng2019large, iqbal2020weakly, xu2022eyes, luo2022cross, huang2023cross}, recent efforts have also delved into spatial~\cite{lucas2020unsup} or temporal~\cite{nyborg2022generalized, capliez2023temporal} domain adaptation for satellite image time series (SITS). However, to the best of our knowledge, no analysis of the impact of temporal or spatial domain shift on the performance of SITS-SCD and of the effects of different design choices has been performed so far. This paper is the first answer to these questions. We leverage the DynamicEarthNet~\cite{Toker_2022_CVPR} and MUDS~\cite{van2021multi} datasets that have both global spatial coverage and multi-year temporal coverage. 

More precisely, we analyze independently the impact of these two domain shifts on both datasets for several methods, giving particular attention to the impact of model size. We evaluate state-of-the-art mono- and bi-temporal semantic segmentation approaches, which process each month or pairs of months independently. We also introduce a multi-temporal SITS-SCD approach that jointly processes images from several months and can leverage long-term temporal information. We show it improves semantic segmentation performance in all settings on both datasets by a significant margin, and that it scales better with model size than the baselines, but that this does not always translate into better change detection performances. We also show that spatial and temporal domain shifts impact SITS-SCD approaches differently, that spatial domain shift has the most dramatic impact, and that while temporal domain shift has limited impact on semantic segmentation performance, it significantly decreases change detection accuracy. In summary, our contributions are as follows:
\begin{itemize}[leftmargin=.3in]
    \item We propose a new architecture to perform direct multi-temporal semantic segmentation that significantly improves SITS-SCD.
    \item We quantify the impact of temporal and spatial shift on the performance of SITS-SCD methods on two global and multi-year SITS datasets for different approaches.
\end{itemize}

\section{Related work}

In this paper, we study satellite image time series semantic change detection (SITS-SCD), considering temporal and spatial domain shifts in our evaluation settings. SITS-SCD consists in segmenting simultaneously each time stamps of a, typically monthly, SITS. We distinguish three strategies to achieve this task. First, \textit{mono-temporal SITS-SCD} methods segment independently each image of the time series. Second, \textit{bi-temporal SITS-SCD} methods segment the input SITS considering image pairs independently. Third, \textit{multi-temporal SITS-SCD} approaches can segment jointly all time stamps of a given SITS. In this section, we review the literature for each of these categories, then list the existing datasets for SITS-SCD, and finally examine how related work tackles potential spatial and temporal shift between training and inference data.

\paragraph{Mono-temporal SITS-SCD} A first set of methods evaluated on SITS-SCD benchmarks predict the semantic segmentation for each time stamps independently. Changes are characterized by the difference of successive semantic maps. These semantic maps can be obtained from a single image using semantic segmentation models like U-Net~\cite{ronneberger2015u}, DeepLabV3~\cite{chen2017rethinking}, Segmenter~\cite{strudel2021segmenter} or Swin-T~\cite{liu2021swin}. If a monthly SITS is to be segmented and multiple images are available for each month, then methods designed for SITS segmentation into a single semantic map, such as 3D-Unet~\cite{m2019semantic}, UTAE~\cite{Garnot_2021_ICCV} or TSViT~\cite{tarasiou2023vits}, can be used as well, using the images from each month independently. \citet{Toker_2022_CVPR} define `monthly', `weekly', and `daily' temporal densities for these approaches, where the first image of the month, six images across the month, or all the images of the month are used respectively to form a monthly SITS. 

\paragraph{Bi-temporal SITS-SCD} These methods perform SITS-SCD for each image pair independently, and are initially designed for the classic semantic change detection (SCD) task that requires to predict the semantic maps of a pair of satellite images at the same location but at distinct time stamps. Very early on, a series of work performed this task following a post-classification procedure~\cite{weismiller1977change, swain1977decision, swain1978bayesian}, where the bi-temporal acquisitions are segmented independently. In this case, the binary change map is obtained as the difference between the predicted segmentation maps. Obviously, such method does not leverage temporal consistencies. To overcome this limitation, another approach is to classify pairs of pixels with transition labels, considering all possible pairs of semantic classes~\cite{bruzzone1997iterative}. However, the number of transition labels increases as the square of the number of semantic classes, and some transitions have very few training examples because there are typically few changes and the land cover classes are very imbalanced (71\%, 11\% and 10\% of Earth's surface is water, forest and agricultural parcels respectively). Such an approach thus faces a very challenging classification setting. Leveraging deep learning advances, most recent approaches \cite{daudt2019multitask, yang2021asymmetric, ding2022bi, zhao2022spatially, xia2022deep, yuan2022transformer, tian2022large, zheng2022changemask, li2023lightweight, jiang2023ttnet, cui2023mtscd, bernhard2023mapformer, ding2024joint, liu2024tbscd} tackle the SCD task with 3-branch models producing two semantic maps and a binary change detection map as output. Multi-task objectives, inner fusion modules in the architecture and/or post-processing operations help guaranty the consistency between the three outputs. In very high resolution, object-based SCD consists in detecting change on identified, often urban, semantic objects (like cars, containers or houses). Objects can be learned as bags of visual words where the dictionary is shared between time stamps~\cite{wu2016scene}, or as temporal correspondences between time stamps~\cite{zhang2017separate}. Our work focuses instead on mid-resolution satellite imagery and generic land cover classes, so object-based SCD is out of the scope of our study. 

\paragraph{Multi-temporal SITS-SCD} Very few methods actually perform SITS-SCD in a multi-temporal manner. \citet{saha2020unsupervised} propose an unsupervised framework for multi-temporal feature learning. Their model processes time stamps independently, the training loss aiming for temporal consistency. While they evaluate their model on classic bi-temporal SCD, one could imagine adapting their method to SITS-SCD but the code is not available. Very close to our approach, TSSCD~\cite{he2024time} is a pixel-wise method extending a one-dimensional fully convolutional network for multi-temporal SITS-SCD. To the best of our knowledge, our proposed method is the first to perform multi-temporal SITS-SCD at the image level.

\paragraph{Semantic change detection datasets} SCD datasets~\cite{daudt2019multitask, yang2021asymmetric, verma2021qfabric, van2021multi, li2022outcome, yuan2022transformer, Toker_2022_CVPR} are intended for the simultaneous semantic segmentation of the land cover at each time stamp and the detection of semantic changes between consecutive time stamps. Datasets like HRSCD~\cite{daudt2019multitask} or SECOND~\cite{yang2021asymmetric} are designed for the bi-temporal task and do not exhibit SITS beyond simple image pairs. The data from the 2021 IEEE GRSS Data Fusion Contest~\cite{li2022outcome} contains time series but only extreme dates have annotations, the challenge focusing more on knowledge transfer from low to high resolution rather than SITS-SCD. QFabric~\cite{verma2021qfabric}, MUDS~\cite{van2021multi} and DynamicEarthNet~\cite{Toker_2022_CVPR} are the SCD datasets the most relevant to our work since they include complete time series and full semantic change annotations. QFabric focuses on urban changes with partial labeling: only changing areas are annotated, with labels such as `Prior Construction', `Land Cleared', or `Construction Done'. Since it is not freely available, we do not consider it. We focus on DynamicEarthNet and MUDS which both contain annotated multi-year SITS covering areas all over the world. DynamicEarthNet is designed for land-use and land-cover classification with classes such as `impervious surface', `forest', or `water', and its areas of interest include a broad range of region types. MUDS, also known as SpaceNet 7, is intended for building tracking over time. We adapt its annotations to the semantic change detection task and propose a first benchmark of SITS-SCD methods on MUDS using the semantic change detection metrics defined by~\citet{Toker_2022_CVPR}.

\paragraph{Temporal and spatial domain shifts} Domain adaptation is a well-known problem with satellite imagery~\cite{deng2019large, iqbal2020weakly, xu2022eyes, luo2022cross, huang2023cross}. In the particular case of SITS, \citet{lucas2020unsup} attempt to adapt state-of-the-art domain adaptation methods to spatial domain shift between two regions within France for the task of generic land cover pixel-wise classification. Crop-type classification with SITS is also relevant for studying temporal domain shift because of seasonal and environmental variability. \citet{capliez2023temporal} examine temporal domain shift in the context of crop type classification in Burkina Faso over multiple years, while \citet{vincent2023pixel} demonstrate the challenges posed by temporal domain shift in agricultural time series pixel-wise classification with a German crop dataset~\cite{kondmann2021denethor}. Additionally, \citet{nyborg2022generalized} propose thermal positional encoding - an encoding based on thermal time rather than calendar time - to account for varying rates of crop growth and mitigate temporal shift issues in Western Europe data. These studies are conducted at the national or continental scale, focus on classification tasks, and mainly try to bridge the performance gap due to domain shift. In contrast, we analyze the impact of temporal and spatial domain shifts independently and at a global scale for our multi-temporal SITS-SCD approach - that outperforms state of the art mono- and bi-temporal approaches - giving particular attention to the impact of model size, an important but often overlooked variable.

\section{Method}

\subsection{Proposed architecture}

\begin{figure*}[t!]
    \centering
    \includegraphics[width=\linewidth]{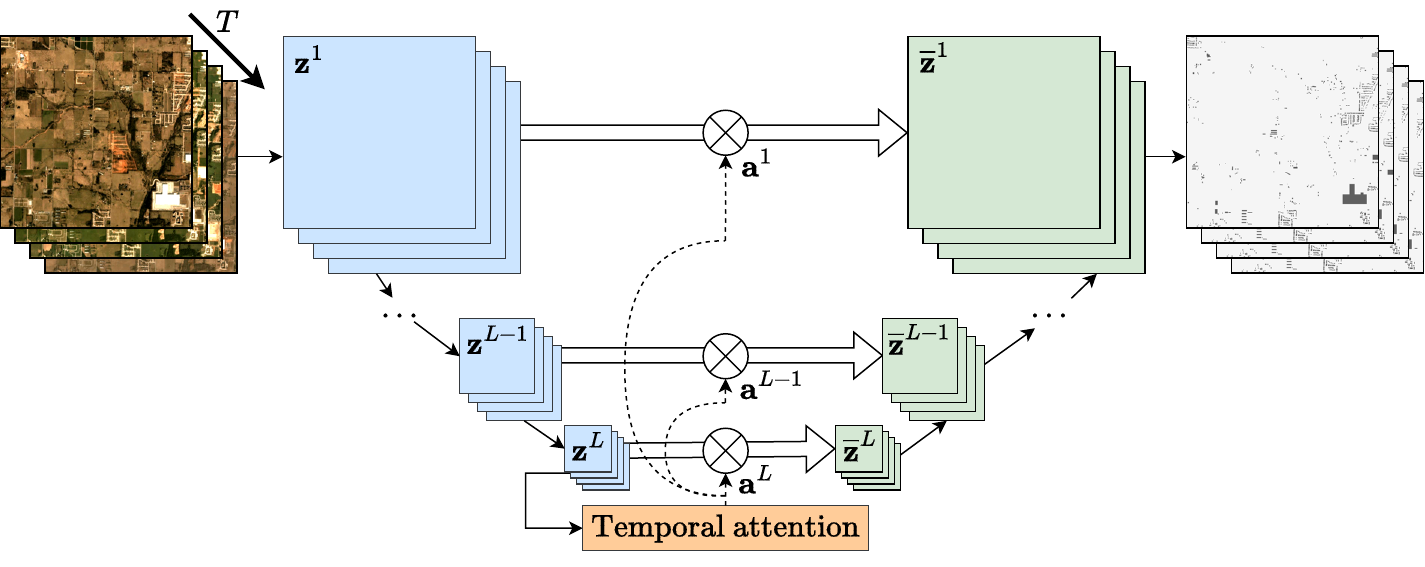}
    \caption{\textbf{Overall architecture.} Given an input SITS, we compute feature maps at various scales. Our contribution is the temporal attention mechanism that allows to account for long-term temporal information. The decoder branch up-scales the feature maps for all time stamps in parallel, while propagating the attention maps at all levels.}
    \label{fig:full_archi}
\end{figure*}

We propose to modify UTAE~\cite{Garnot_2021_ICCV} by changing the core temporal attention mechanism to output one segmentation map per input image instead of aggregating temporal information, in order to better leverage temporal knowledge. The overall pipeline is illustrated by Figure~\ref{fig:full_archi} where we show the encoder branch, the temporal attention block, and the decoder branch.

\paragraph{Encoder} Our encoder takes as input a SITS $\inputseq$ in $\mathbb{R}^{T\times H\times W\times C}$ of $T$ satellite images of spatial dimensions $H\times W$ with $C$ spectral bands. The encoder branch of our model strictly mirrors UTAE and produces a series of feature maps $\fmap^1$, ..., $\fmap^L$ using $L$ successive down-sampled convolutions. Positional encoding is added to the feature map at the last level $\fmap^L$ in $\mathbb{R}^{T\times H'\times W'\times D}$, with $H'\times W'$ the spatial resolution and $D$ the feature size at level $L$. Similar to~\citet{Garnot_2021_ICCV}, sections of size $D/h$ of $\fmap^L$ are processed independently in a $h$-head manner. For the sake of conciseness, we ignore positional encoding and multi-head processing in our notations in the following sections.

\begin{figure}
    \centering
    \resizebox{\linewidth}{!}{
    \begin{tabular}{c}
         \vspace{-0.7em}
         \includegraphics[trim={2cm 0 0 0}, clip, width=\linewidth]{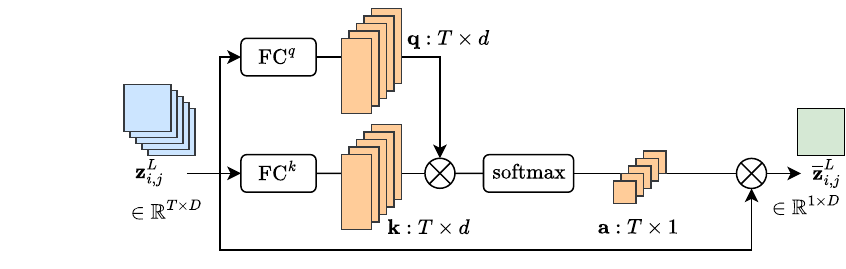}\\
         (a) TAE~\cite{Garnot_2020_CVPR}\\
         \vspace{-0.7em}
         \includegraphics[trim={2cm 0 0 0}, clip, width=\linewidth]{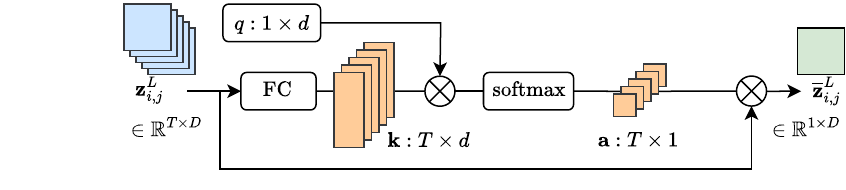}\\
         \vspace{+0.3em}
         (b) LTAE~\cite{Garnot_2020_AALTD}\\
         \vspace{-0.7em}
         \includegraphics[trim={2cm 0 0 0}, clip, width=\linewidth]{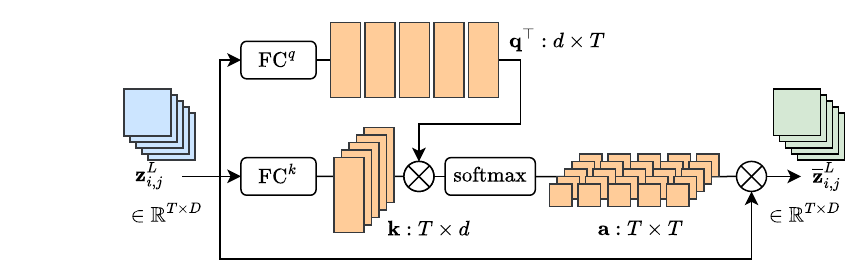}\\
         (c) Ours
    \end{tabular}
    }
    \caption{\textbf{Attention mechanism of TAE, LTAE and our method.} We show the temporal attention mechanism of TAE~\cite{Garnot_2020_CVPR}, LTAE~\cite{Garnot_2020_AALTD} and our method for a given patch $\fmap^L_{i,j}$ of the feature map $\fmap^L$. Here, $d$ is the dimension of the key and query vectors.}
    \label{fig:temp_archi}
\end{figure}

\paragraph{Attention mechanism} Our temporal attention mechanism outputs multi-temporal attention maps $\attention^L$ in $[0,1]^{T\times T\times H'\times W'}$ at the lowest resolution. Its role is to combine the different temporal feature maps while maintaining a temporal dimension. For each time stamp $t$ in the range $\{1, \ldots, T\}$, we aim to incorporate information from all dates into the prediction for $t$, with the contributions of each date $t'$ in the range $\{1, \ldots, T\}$ being specific to $t$. Our proposed attention mechanism builds on TAE~\cite{Garnot_2020_CVPR}, which predicts the queries as a function of the feature maps at the lowest level $\fmap^L$. However, instead of computing the attention weights as a scalar product of the keys and queries, we define the weights as their matrix multiplication in order to keep the temporal dimension. Note that UTAE, on which our overall architecture is built upon, uses a lightweight temporal attention encoder (LTAE)~\cite{Garnot_2020_AALTD} at its core to aggregate the temporal feature maps. LTAE is a lightweight version of TAE where the queries are free parameters of the model and are the same for all time stamps. We illustrate in Figure~\ref{fig:temp_archi} the differences between TAE, LTAE and our proposed attention mechanism. For a given spatial location $(i,j)$ in $[1,H']\times[1,W']$, we compute the attention weights $\attention_{i,j}^L$ from queries
\begin{equation}
    \mathbf{q} = \text{FC}^q(\fmap^L_{i,j}) \ \in\mathbb{R}^{T\times d},
    \label{eq:queries}
\end{equation}
and keys
\begin{equation}
    \mathbf{k} = \text{FC}^k(\fmap^L_{i,j}) \ \in\mathbb{R}^{T\times d},
    \label{eq:keys}
\end{equation}
as
\begin{equation}
    \attention_{i,j}^L = \mathbf{k}\mathbf{q}^\top\ \in\mathbb{R}^{T\times T},
    \label{eq:attention_maps}
\end{equation}
where $\text{FC}^q$ and $\text{FC}^k$ denote fully-connected layers. The attention maps are up-sampled at all levels $l$ in $\{1, \ldots, L\}$ into attention maps $\attention^l$, so that the combined feature map is obtained for all time stamps $t$ as:
\begin{equation}
    \overline{\fmap}_t^l = \sum_{t'=1}^T \attention_{t,t'}^l \odot\fmap_{t'}^l,
    \label{eq:attention}
\end{equation}
where $\odot$ denotes the element-wise multiplication.

\paragraph{Decoder} The decoder uses strided transposed convolutions to up-sample the feature maps $\overline{\fmap}^l$ to the upper level. Following~\citet{Garnot_2021_ICCV}, we propagate the up-sampled attention maps $\attention^1, \ldots, \attention^{L-1}$ at all levels with skip connections: before each up-sampling convolution, the obtained feature map $\overline{\fmap}_t^l$ is concatenated to the up-sampled feature map of the lower level $\overline{\fmap}_t^{l+1}$. All time steps are processed in parallel using the same decoder branch. Outputs are the segmentation maps $\segmap$ in $\mathbb{R}^{T\times H\times W\times K}$ where $K$ is the number of semantic classes. 

\subsection{Analysis methodology}\label{sec:dom_setup}

\begin{figure}[t!]
\centering
\begin{tabular}{c}
    \includegraphics[width=0.98\linewidth]{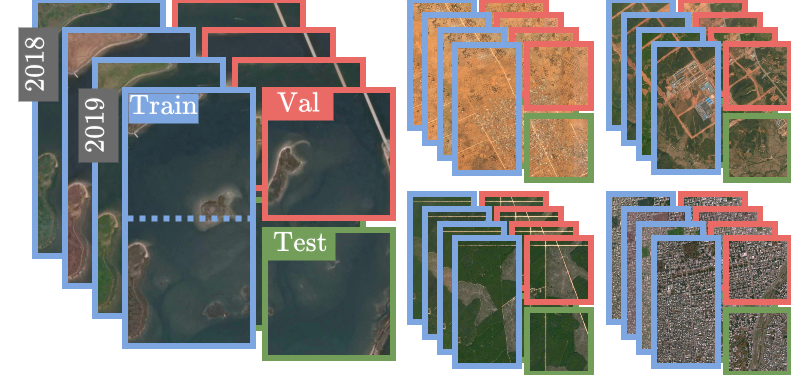}\vspace{-.5em}\\
    (a) No domain shift setting\vspace{.7em}\\
    \includegraphics[width=0.98\linewidth]{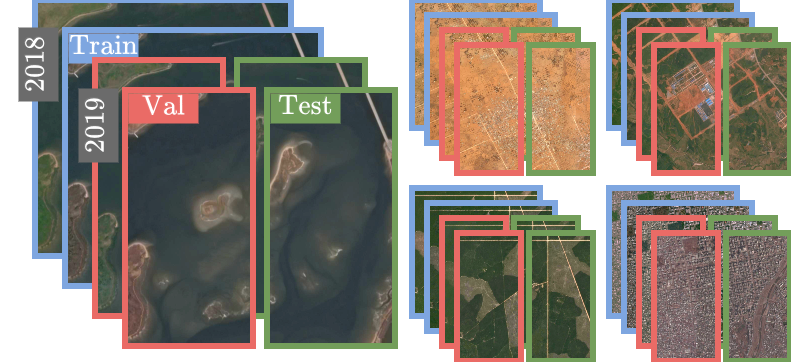}\vspace{-.5em}\\
    (b) Temporal domain shift setting\vspace{.7em}\\
    \includegraphics[width=0.98\linewidth]{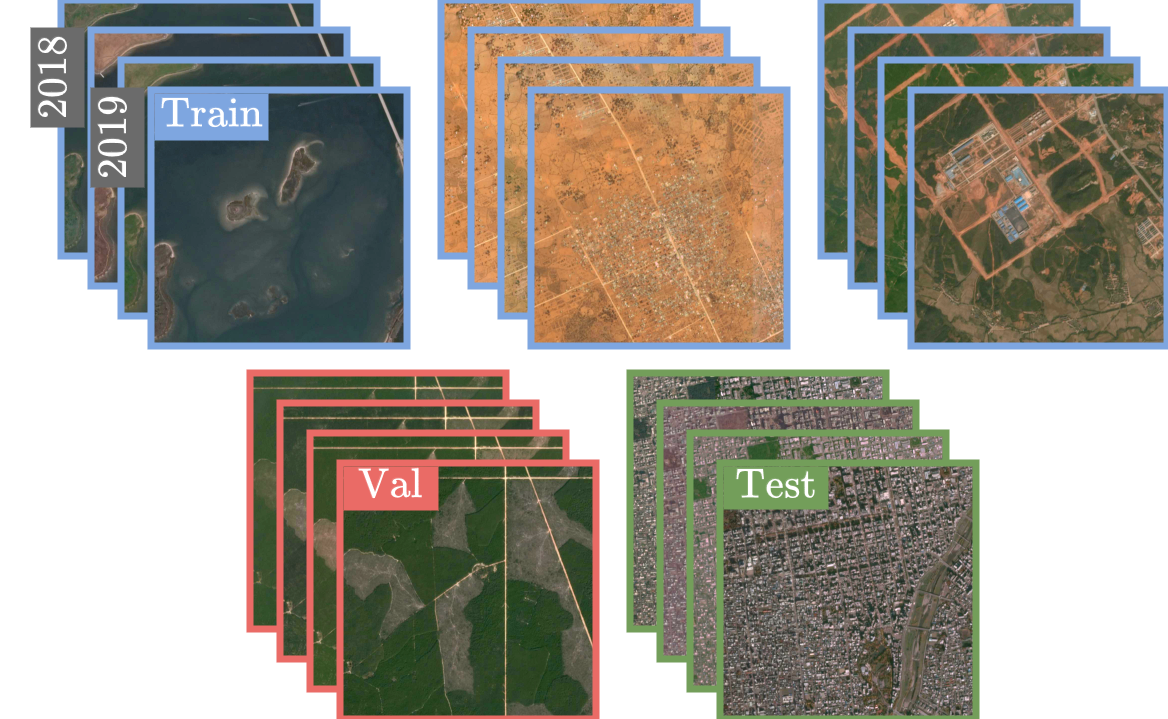}\vspace{-.5em}\\
    (c) Spatial domain shift setting
    \end{tabular}
    \caption{\textbf{Domain shift settings.} We organize the dataset splits in three different manners such that there is respectively (a) no domain shift, (b) a temporal domain shift, and (c) a spatial domain shift between train and val/test sets. DynamicEarthNet~\cite{Toker_2022_CVPR} images are shown here for visualization, and we use the same settings for MUDS~\cite{van2021multi}.}\vspace{-.5em}
\label{fig:folds}
\end{figure}

We perform our analysis on two global and multi-year SITS datasets: DynamicEarthNet~\cite{Toker_2022_CVPR} and MUDS~\cite{van2021multi}. We evaluate all methods in three different settings: a setting without domain shift between train and test sets, a setting with temporal shift, and a setting with spatial domain shift. These different settings are visualized in Figure~\ref{fig:folds}.

\paragraph{No domain shift} We split the datasets into 4 subsets as visualized in Figure~\hyperref[fig:folds]{\ref*{fig:folds}a}: we split each SITS into four SITS of equal size. We keep two for training, one for validation and one for test purposes. We follow a 4-fold cross validation scheme that we detail in Section~\hyperref[sec:suppmatdata]{A} of the appendix. Though folds cover distinct areas, they all share common regions so that there is no significant spatial domain shift. Additionally, since all splits cover the same two years, there is thus no temporal shift.

\paragraph{Temporal shift} DynamicEarthNet and MUDS contain 2-year image time series from January 2018 to December 2019. We split all time series in half, keeping 2018 for training and 2019 for validation and test purposes as visualized in Figure~\hyperref[fig:folds]{\ref*{fig:folds}b}. Note that more annual data would be necessary to create multiple folds for cross-validation in the temporal setting. In this setting, there is no spatial domain shift.

\paragraph{Spatial shift} We split the 55 (resp. 60) areas of interest of DynamicEarthNet (resp. MUDS) into five subsets of 11 (resp. 12) image time series. As illustrated in Figure~\hyperref[fig:folds]{\ref*{fig:folds}c}, we keep three sets for training while the remaining two are kept for validation and test purposes respectively. We then follow a 5-fold cross validation scheme described in Section~\hyperref[sec:suppmatdata]{A} of the appendix. Here, all subsets cover regions that are significantly different and far from each other so that there is a spatial domain shift. Folds are random on MUDS and selected so that the class distribution is approximately similar in each fold on DynamicEarthNet. More details to reproduce this setting can be found in Section~\hyperref[sec:suppmatdata]{A} of the appendix. Note that this spatial shift setting is the one used in the DynamicEarthNet\footnote{https://codalab.lisn.upsaclay.fr/competitions/2882} and the SpaceNet 7\footnote{https://spacenet.ai/sn7-challenge/} challenges.

\subsection{Training and implementation details}

Following~\citet{tarasiou2023vits}, we trained all methods using focal loss~\cite{lin2017focal} and the AdamW optimizer~\cite{loshchilov2017decoupled} with a learning rate starting from 0 and gradually reaching 10$^{-4}$ as a warmup after 5000 iterations. We train our model for 500~000 iterations and keep the checkpoint that achieves the best semantic change segmentation score on the validation set. For data augmentation, we randomly crop image patches of size 128$\times$128 out of the 1024$\times$1024 images and additionally do random horizontal and vertical flips as well as random rotations. 

Our model can take image sequences of variable length as input. During training, we sample 12 random monthly images out of the 24 available in the setting without domain shift and the spatial shift setting. In the temporal shift setting, we sample 6 out of the 12 available images. At inference, we take the full monthly time series as input, \textit{i.e.} a sequence of length 24 or 12 depending on the case. In Section~\ref{sec:leveraging}, we study alternative inference schemes.

We set as default values for the number of levels $L=4$, for the spatial feature size $D=512$ and for the dimension of keys and queries $d=4$. We investigate the impact of changing $D$ in Section~\ref{sec:leveraging} and $d$ in Section~\hyperref[sec:impactofd]{C} of the appendix.

\section{Experiments}

\subsection{Datasets and metrics}

\input{tables/results_all_setups}
\begin{table}[t!]
  \renewcommand{\arraystretch}{0.9}
  \setlength\tabcolsep{3pt}
  \centering
  \resizebox{\linewidth}{!}{
  \begin{tabular}{ccccccccccccccccccccccccccccc}
  \toprule
    Seq. & \multicolumn{24}{c}{\multirow{2}{*}{Inference time series splitting}} & \multirow{2}{*}{SCS$\uparrow$} & \multirow{2}{*}{SC$\uparrow$} & \multirow{2}{*}{BC$\uparrow$} & \multirow{2}{*}{mIoU$\uparrow$}\\
    len. & & & & & & & & & & & & & & & & & & & & & & & & & & & & \\
    \midrule
    \multirow{2}{*}{6} 
    &\scriptsize\bc & \bc & \bc & \bc & \bc & \bc & \rc & \rc & \rc & \rc & \rc & \rc & \gc & \gc & \gc & \gc & \gc & \gc & \kc & \kc & \kc & \kc & \kc & \kc & 27.9 & 38.0 & 17.8 & 56.9\\
    &\bc & \rc & \gc & \kc & \bc & \rc & \gc & \kc & \bc & \rc & \gc & \kc & \bc & \rc & \gc & \kc & \bc & \rc & \gc & \kc & \bc & \rc & \gc & \kc & 29.0 & \textbf{41.5} & 16.6 & 59.5\\
    \midrule
    \multirow{2}{*}{8} 
    &\bc & \bc & \bc & \bc & \bc & \bc & \bc & \bc & \rc & \rc & \rc & \rc & \rc & \rc & \rc & \rc & \gc & \gc & \gc & \gc & \gc & \gc & \gc & \gc & 29.7 & 39.6 & 19.8 & 58.3\\
    &\bc & \rc & \gc & \bc & \rc & \gc & \bc & \rc & \gc & \bc & \rc & \gc & \bc & \rc & \gc & \bc & \rc & \gc & \bc & \rc & \gc & \bc & \rc & \gc & 29.3 & 41.2 & 17.3 & 59.9\\
    \midrule
    \multirow{6}{*}{12} 
    &\bc & \bc & \bc & \bc & \bc & \bc & \bc & \bc & \bc & \bc & \bc & \bc & \gc & \gc & \gc & \gc & \gc & \gc & \gc & \gc & \gc & \gc & \gc & \gc & 30.6 & 40.1 & 21.1 & 59.6\\
    &\bc & \bc & \bc & \bc & \bc & \bc & \gc & \gc & \gc & \gc & \gc & \gc & \bc & \bc & \bc & \bc & \bc & \bc & \gc & \gc & \gc & \gc & \gc & \gc & 29.8 & 39.7 & 19.9 & 59.0\\
    &\bc & \bc & \bc & \bc & \gc & \gc & \gc & \gc & \bc & \bc & \bc & \bc & \gc & \gc & \gc & \gc & \bc & \bc & \bc & \bc & \gc & \gc & \gc & \gc & 30.9 & 41.2 & 20.6 & 59.9\\
    &\bc & \bc & \bc & \gc & \gc & \gc & \bc & \bc & \bc & \gc & \gc & \gc & \bc & \bc & \bc & \gc & \gc & \gc & \bc & \bc & \bc & \gc & \gc & \gc & 30.4 & 40.8 & 20.0 & 59.8\\
    &\bc & \bc & \gc & \gc & \bc & \bc & \gc & \gc & \bc & \bc & \gc & \gc & \bc & \bc & \gc & \gc & \bc & \bc & \gc & \gc & \bc & \bc & \gc & \gc & 30.4 & 41.3 & 19.5 & 60.1\\
    &\bc & \gc & \bc & \gc & \bc & \gc & \bc & \gc & \bc & \gc & \bc & \gc & \bc & \gc & \bc & \gc & \bc & \gc & \bc & \gc & \bc & \gc & \bc & \gc & 29.8 & 41.3 & 18.3 & 60.2\\
    \midrule
    24 & \bc & \bc & \bc & \bc & \bc & \bc & \bc & \bc & \bc & \bc & \bc & \bc & \bc & \bc & \bc & \bc & \bc & \bc & \bc & \bc & \bc & \bc & \bc & \bc & \textbf{31.7} & 41.0 & \textbf{22.4} & \textbf{60.5}\\
    \bottomrule
    \multicolumn{29}{c}{\includegraphics[width=1.1\linewidth]{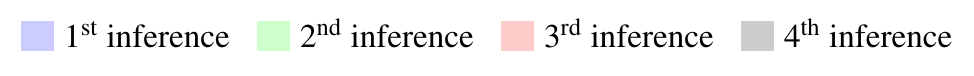}}\vspace{-1em}
  \end{tabular}
  }
  \caption{\textbf{Inference time series size.} We report the SCS, SC, BC and mIoU of our model for various input sequence length and different splitting configurations. The splitting is best viewed in color, where cells of the same color were gathered together as a SITS to produce their corresponding predictions simultaneously.}\vspace{-1em}
  \label{tab:scores_ts_sizes}
\end{table}

We evaluate our method along with baselines on DynamicEarthNet~\cite{Toker_2022_CVPR} and MUDS~\cite{van2021multi} datasets and more precisely on their training set for which ground truth annotations are available. Images for both these datasets were acquired by Planet Labs with a ground sample distance (GSD) of approximately 3 meters.

\paragraph{DynamicEarthNet~\cite{Toker_2022_CVPR}} This dataset contains 55 daily SITS from January, 1st 2018 to December, 31st 2019 distributed over the globe. The first day of each month is annotated, leading to 24 ground truth segmentation maps per area of interest (AoI). Images are of size 1024$\times$1024 and multi-spectral with 4 channels (RGB + near-infrared). Annotations are general land-use and land-cover classes: `impervious surface', `agriculture', `forest', `wetlands', `soil' and `water'. The `snow' class is only present on very few AoIs of the dataset and is discarded for this study.

\paragraph{MUDS~\cite{van2021multi}} The Multi-temporal Urban Development SpaceNet (MUDS) dataset consists of 60 monthly SITS collected between 2017 and 2020 all over the globe. MUDS contains few images acquired in 2017 and 2020: they are discarded for this study. Due to an excessive amount of clouds or haze some images were excluded from the dataset, causing a few gaps in some of the time series, with length ranging from 18 to 24 images per SITS. Images are of size 1024$\times$1024 with 3 channels (RGB). Default annotations for this dataset are polygons indicating buildings from which we generate binary segmentation maps with classes `building' and `not building'.

\paragraph{Metrics}

We use the four metrics defined by~\citet{Toker_2022_CVPR} to assess the performance of evaluated methods. The mean intersection-over-union (mIoU) on the semantic labels indicates the ability of the method to predict correct semantic segmentations, irrespective of the change. The binary change (BC) score depicts how well a method can predict a semantic change while the semantic change (SC) score focuses on the semantic prediction for pixels where a change actually occurs. The semantic change segmentation (SCS) score is the average of both previous scores.

\subsection{Baselines}

\subsubsection{Mono-temporal}

We evaluate two state-of-the-art semantic segmentation methods designed for SITS: UTAE~\cite{Garnot_2021_ICCV} and TSViT~\cite{tarasiou2023vits}. 

\paragraph{UTAE~\cite{Garnot_2021_ICCV}} The U-Net with Temporal Attention Encoder (UTAE) consists of a U-Net architecture where a temporal attention mechanism squeezes the temporal dimension before the decoding branch. Thus, the model outputs a single segmentation map for the whole input time series. This method shows competitive performance on recent segmentation benchmarks~\cite{Toker_2022_CVPR, tarasiou2023vits}. 

\paragraph{TSViT~\cite{tarasiou2023vits}} In contrast to UTAE, the Temporo-Spatial Vision
Transformer (TSViT) has a fully-attentional architecture, processing the tokens first temporally then spatially. It was shown to improve semantic segmentation on several datasets~\cite{tarasiou2023vits}.\\

Both these methods output a single prediction map for a given SITS as input. In order to evaluate them on the SITS-SCD task, we follow the setting of~\citet{Toker_2022_CVPR} where the monthly segmentation maps are predicted independently from one another by using as input signal one or several images in the month. In the \textit{monthly} setting, only the first image of each month is used, and the input time series is actually composed of a single image. In the \textit{weekly} setting, a SITS of six images - corresponding to an image every 5 days through the month - serves as input to obtain the monthly prediction. \citet{Toker_2022_CVPR} show that a \textit{daily} setting - where all images of a month are used as a SITS to predict a monthly segmentation map - does not improve over the weekly setting, thus we do not consider it in this work. We set same values of $L$, $D$ and $d$ for UTAE-based methods as with our method for fair comparison. For TSViT, we set the feature dimension to 512 so that the number of trainable parameters is of the same order of magnitude as other evaluated methods. Additional details on our UTAE and TSViT implementations are provided in Section~\hyperref[sec:impl_details]{B} of the appendix.  

\subsubsection{Bi-temporal}
We evaluate two state-of-the-art bi-temporal SCD methods: A2Net~\cite{li2023lightweight} and SCanNet~\cite{ding2024joint}.

\paragraph{A2Net~\cite{li2023lightweight}} A2Net first extracts multi-stage feature maps from a pair of images with a shared-weight MobileNetV2~\cite{sandler2018mobilenetv2}. The difference of the two feature maps at all stages are combined and decoded into a binary change mask and two semantic segmentation maps.

\paragraph{SCanNet~\cite{ding2024joint}} The Semantic
Change Network (SCanNet) has a three-branch encoder-decoder architecture. The image pairs and the concatenation of intermediate feature representations are used to learn two sets of semantic tokens (one for each time stamps) and a set of change tokens. All tokens are concatenated and processed by an inner transformer. The output is decoded into a binary change mask and two semantic segmentation maps.\\

There are multiple manners to adapt bi-temporal methods to SITS-SCD. We train both methods with all possible ordered image pairs. At inference, a SITS is divided in consecutive image pairs that are segmented independently. We only consider the semantic outputs and disregard the predicted binary change maps at inference. Additional details on our A2Net and SCanNet implementations are provided in Section~\hyperref[sec:impl_details]{B} of the appendix.  

\subsubsection{Multi-temporal}

\noindent We evaluated {TSSCD~\cite{he2024time}}, a state-of-the-art pixel-wise multi-temporal approach, adapting a one-dimensional fully convolutional network to the SITS-SCD task. Its training requires a particular sampling of pixel time series that we detail in Section~\hyperref[sec:impl_details]{B} of the appendix.  
\subsection{Results}

\begin{figure}[t!]
\centering
\setlength{\tabcolsep}{2pt}
\begin{tabular}{cccc}
    \includegraphics[width=0.49\linewidth]{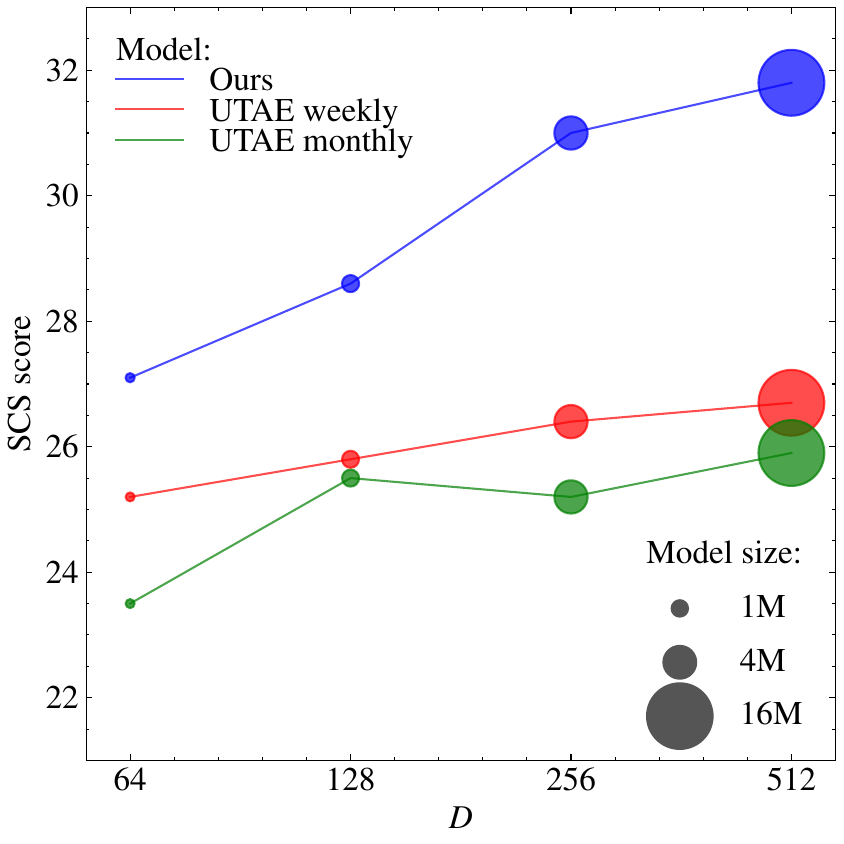} & \includegraphics[width=0.49\linewidth]{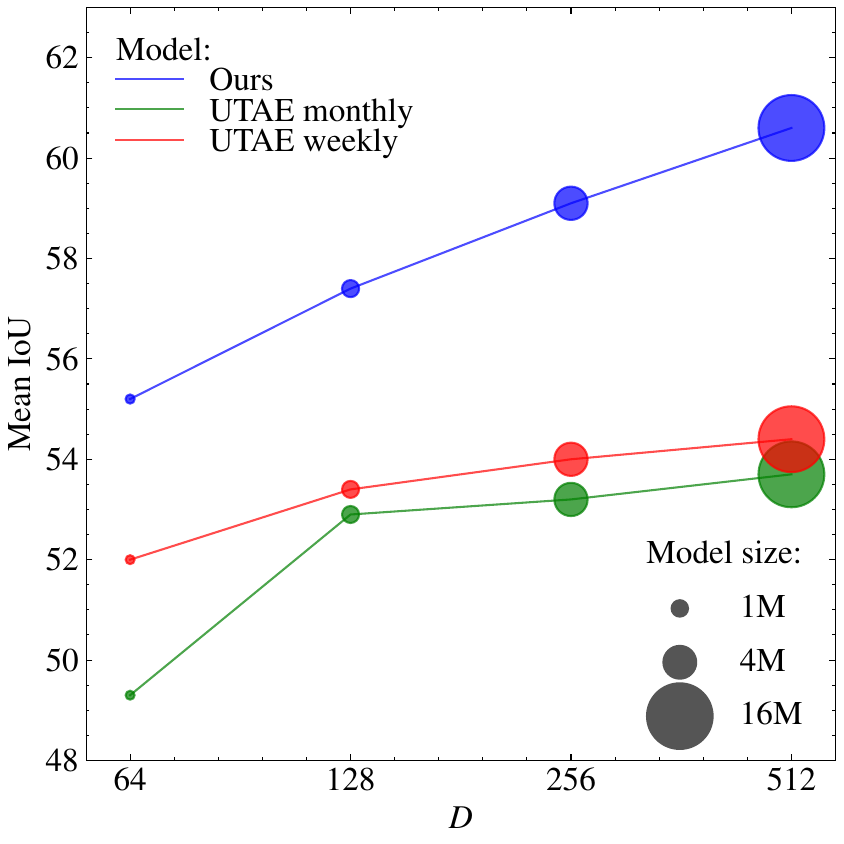}\vspace{-.5em}\\
    \multicolumn{2}{c}{(a) Method comparison}\vspace{.5em}\\
    \includegraphics[width=0.49\linewidth]{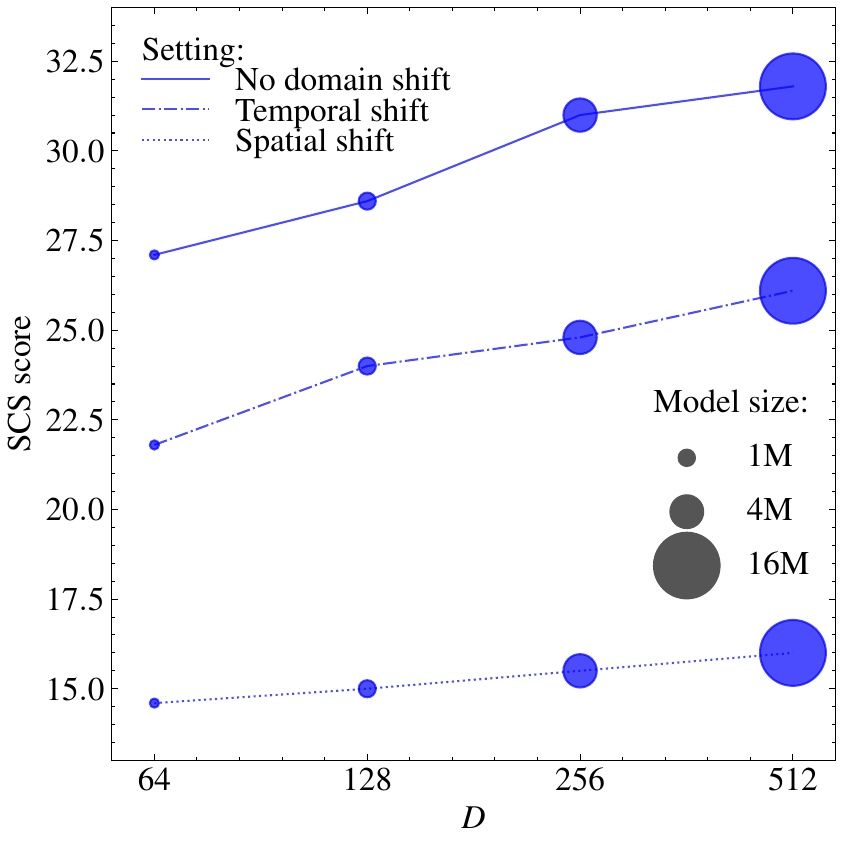} & \includegraphics[width=0.49\linewidth]{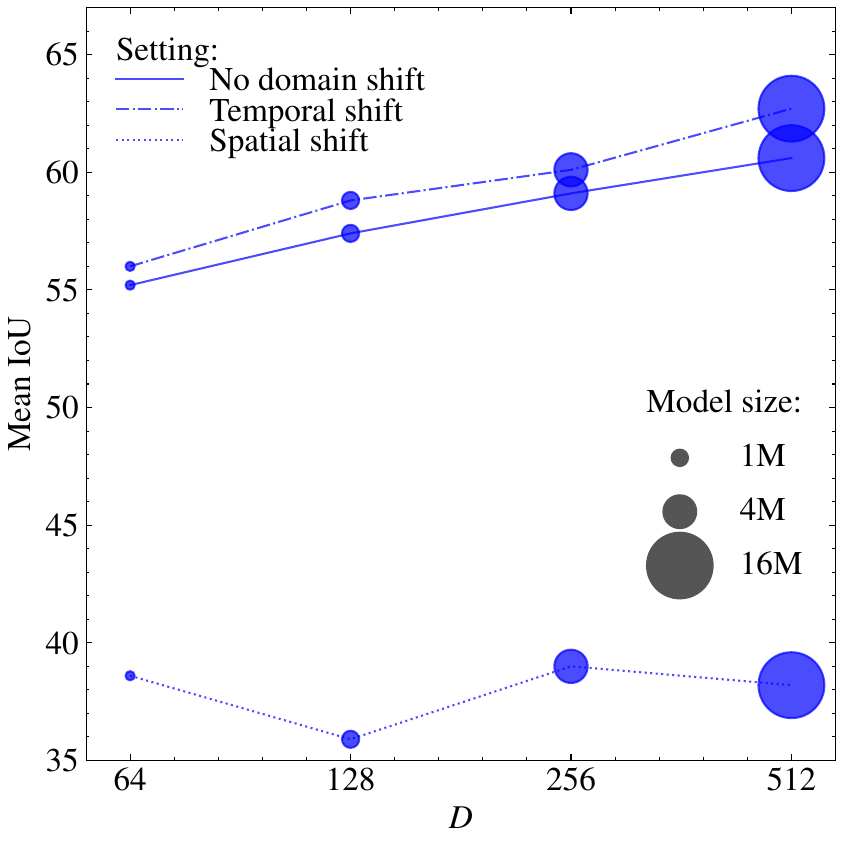}\vspace{-.5em}\\
    \multicolumn{2}{c}{(b) Setting comparison}\\
\end{tabular}
\caption{\textbf{Impact of $D$ on performance.} We compare the impact of the spatial feature size $D$ on performance (a) for our model and UTAE-based methods in the setting without domain shift and (b) for our model in the three domain shift settings. In each case, we report the SCS score (left) and the mean IoU (right).}\vspace{-.5em}
\label{fig:spat_temp_feat}
\end{figure}
\begin{figure*}[t!]
\centering
\setlength{\tabcolsep}{1pt}
\resizebox{\linewidth}{!}{
    \begin{tabular}{cccccccccccc}
        & {\scriptsize (i)} & {\scriptsize (ii)} & {\scriptsize (iii)} & {\scriptsize (iv)} & {\scriptsize (v)} & {\scriptsize (vi)} & {\scriptsize (vii)} & {\scriptsize (viii)} & {\scriptsize (ix)} & {\scriptsize (x)} & {\scriptsize (xi)}\\ 
        \rotatebox[origin=c]{90}{\scriptsize T1} & \satimgchange{inputchange1}{}\vspace{2pt}\\
        \rotatebox[origin=c]{90}{\scriptsize T2} & \satimgchange{inputchange2}{}\vspace{2pt}\\
        \rotatebox[origin=c]{90}{\scriptsize GT} & \satimgchange{gtchange}{}\vspace{2pt}\\
        \rotatebox[origin=c]{90}{\scriptsize Ours} & \satimgchange{ours}{_colors}\vspace{2pt}\\
        \rotatebox[origin=c]{90}{\scriptsize TSViT} & \satimgchange{tsvit}{_colors}\vspace{2pt}\\
        \rotatebox[origin=c]{90}{\scriptsize UTAE} & \satimgchange{utae}{_colors}\vspace{2pt}\\
        \rotatebox[origin=c]{90}{\scriptsize TSSCD} & \satimgchange{tsscd}{}\vspace{2pt}\\
        \rotatebox[origin=c]{90}{\scriptsize SCanNet} & \satimgchange{scannet}{}\vspace{2pt}\\
        \rotatebox[origin=c]{90}{\scriptsize A2Net} & \satimgchange{a2net}{}\\
        \multicolumn{12}{c}{\includegraphics[width=0.68\linewidth]{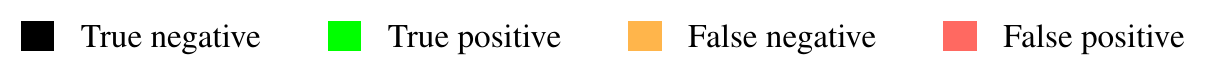}}
    \end{tabular}
    }\vspace{-.9em}
    \caption{{\bf Qualitative change detection results.} We show the binary change detection maps predicted by our model and competing methods in the setting without domain shift for randomly selected input images. From top to bottom, we show the input pairs at time T1 (01/09/2018) and T2 (01/09/2019), the ground truth binary change map and the predictions of different methods for DynamicEarthNet  (i-vi) and MUDS (vii-xi). For TSViT and UTAE we use the weekly setting for DynamicEarthNet and the monthly setting for MUDS.  Best viewed in color.}\vspace{-.6em}
    \label{fig:method_qualitative}
\end{figure*}
\begin{figure*}[t!]
\centering
\setlength{\tabcolsep}{1pt}
\resizebox{\linewidth}{!}{
    \begin{tabular}{ccccccccccccc}
        & & {\scriptsize (i)} & {\scriptsize (ii)} & {\scriptsize (iii)} & {\scriptsize (iv)} & {\scriptsize (v)} & {\scriptsize (vi)} & {\scriptsize (vii)} & {\scriptsize (viii)} & {\scriptsize (ix)} & {\scriptsize (x)} & {\scriptsize (xi)}\\ 
        \multicolumn{2}{c}{\rotatebox[origin=c]{90}{\scriptsize Input}} & \satimgsetup{input}\vspace{2pt}\\
        \multicolumn{2}{c}{\rotatebox[origin=c]{90}{\scriptsize GT}} & \satimgsetup{gt}\vspace{2pt}\\
        \rotatebox[origin=c]{90}{\scriptsize No do-}&\rotatebox[origin=c]{90}{\scriptsize main shift} & \satimgsetup{default}\vspace{2pt}\\
        \rotatebox[origin=c]{90}{\scriptsize Temporal}&\rotatebox[origin=c]{90}{\scriptsize shift} & \satimgsetup{temporal}\vspace{2pt}\\
        \rotatebox[origin=c]{90}{\scriptsize Spatial}&\rotatebox[origin=c]{90}{\scriptsize shift} & \satimgsetup{spatial}\\    
        \multicolumn{9}{c}{\includegraphics[width=0.68\linewidth]{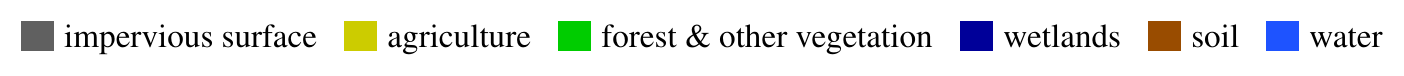}}& \multicolumn{4}{c}{\includegraphics[width=0.205\linewidth]{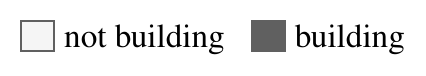}}
    \end{tabular}
    }\vspace{-.7em}
    \caption{{\bf Qualitative segmentation results in different settings.} We show the segmentation maps predicted by our model in different settings for randomly selected input images. From top to bottom, we show the input image (on 01/07/2019), the corresponding ground truth, the predictions without domain shift, the predictions with temporal shift and the predictions with spatial shift. Images from (i-vii) are taken from DynamicEarthNet and (viii-xi) from MUDS. We highlight areas where our method fails in the spatial domain shift setting on the {\color{red}`agriculture'} {\color{red}$\mathbf{\bigcirc}$} and the {\color{RubineRed}`impervious surface'} {\color{RubineRed}$\mathbf{\bigcirc}$} classes for DynamicEarthNet and the {\color{blue}`building class'} {\color{blue}$\mathbf{\bigcirc}$} for MUDS. Best viewed in color.}\vspace{-.6em}
    \label{fig:setup_qualitative}
\end{figure*}

\subsubsection{Leveraging long-term temporal information}\label{sec:leveraging}

In Table~\ref{tab:scores_all_setups}, we report the performance obtained on DynamicEarthNet and MUDS in all three settings. Additional quantitative and qualitative results can be found in Sections~\hyperref[sec:suppmatquant]{D} \&~\hyperref[sec:suppmatqual]{E} of the appendix. Our architecture performs better than other evaluated baselines in all settings and on both datasets in terms of binary classification and semantic segmentation. We discuss SC and SCS performance in Section~\ref{sec:limits}. We explain the better scores obtained by our model by its ability to extract temporal knowledge from long-term images of the time series. 

To confirm this intuition, we investigate various inference schemes in the setting without domain shift and report the results in Table~\ref{tab:scores_ts_sizes}. We evaluate the performance of our method when performing inference on sub-sequences of the full 24-month image sequence of sizes 6, 8, 12 and 24. We also explore various ways to sample these sub-sequences, as visualized by the colors on the left of the Table. The results clearly show that the performance improves when longer series are used at inference, validating our hypothesis that our model leverages temporal information over a long temporal range. Interestingly, one can also see that for a given input sequence length, having sub-sequences that span the 24-month period is better than using successive slices of the full time series. This seems particularly important when using short sequences.

Another indicator of our method's ability to learn more informative features than UTAE-based methods is that its performance improves as the spatial feature size $D$ increases, as illustrated on Figure~\hyperref[fig:spat_temp_feat]{\ref*{fig:spat_temp_feat}a}. While the SCS score and the mIoU of our architecture increases as the model gets bigger, there is no significant increase in UTAE performance for feature size higher than $D=128$. This shows our architecture can better leverage spatial information.

Qualitatively, we report the predicted binary change detection maps for pairs of images one year apart without domain shift in Figure~\ref{fig:method_qualitative}. Our predictions show significantly fewer false positive than competing baselines.

\subsubsection{Comparing domain shift settings}

The results from Table~\ref{tab:scores_all_setups} show that spatial domain shift has the most significant impact on performance, both regarding semantic and change detection scores. For all evaluated methods, we observe an average relative drop of the mIoU from the setting without domain shift to the spatial shift setting of 31.9\% on DynamicEarthNet and of 10.5\% on MUDS. The SCS score similarly decreases by 39.0\% on DynamicEarthNet and by 20.7\% on MUDS. This drop is explained by the diversity of geographies contained in these two global datasets. The fact the performance drop is smaller on MUDS than on DynamicEarthNet is likely related to the fact that the geographic variability is less pronounced on buildings than on other land-cover types. On DynamicEarthNet, the drop of IoU for the `impervious surface' class (\textit{i.e.} artificial land) is only of 23.0\%, similar to the drop observed on MUDS, while it is 49.7\% on average for the other land-cover classes.

The impact of the domain shift can be seen qualitatively in Figure~\ref{fig:setup_qualitative}, where we show segmentation results of our method trained in each setting on the same images. We highlight in {\color{red}red circles} areas where `agriculture' is classified as `forest' and in {\color{RubineRed}pink circles}, areas where `impervious surface' is classified as `soil' in the spatial setting on DynamicEarthNet. On MUDS, though some buildings are not detected in the spatial setting as highlighted by the {\color{blue}blue circles}, our method seems to rarely classify `not building' as `building' in any of the settings.

In figure~\hyperref[fig:spat_temp_feat]{\ref*{fig:spat_temp_feat}b}, we analyze the relation between the model size and performance for our method in the different domain shift settings. Two effects are striking. First, while performance exhibits gradual improvement as the number of parameters increases in the absence of domain shift and under temporal shift conditions, there is no significant improvement in mIoU in the spatial shift setting, and only a slight increase in the SCS score. This again highlights the importance of spatial domain shift. Second, this graph confirms that the semantic segmentation results are similar without domain shift and in the temporal setting (which can also be seen qualitatively in Figure~\ref{fig:setup_qualitative}), but change detection performance is clearly impacted by temporal domain shift. We believe we are the first to highlight this very specific impact of temporal domain shift for change detection.

\subsubsection{Limits of current methods and future work}\label{sec:limits}

Binary change detection on DynamicEarthNet and MUDS is a challenging task. The BC score in all settings and for all methods is relatively low, below 23\%. This clearly is an obstacle for the application of current methods for SITS-SCD. To understand this low performance, it is important to note that there are very few changes occurring in these two datasets: the proportion of pixels that semantically change from one month to the next is of 1.28\% in DynamicEarthNet and of 0.03\% in MUDS. This makes the SC and SCS scores, for which our method is not always better than competing baselines, hard to interpret since they focus on very few pixels on which change is accurately detected. However, this is consistent with practical use cases where semantic changes, \textit{e.g.} the construction/destruction of a building, droughts/floods or deforestation, are rare events at a global scale. Thus, while we believe that adapting SITS-SCD methods to address temporal and spatial domain shifts for multi-year and global applications is an important challenge, we also argue that improving performance in the setting without any domain shift is important, and will likely require better addressing the scarcity of change data. On MUDS especially, no method significantly outperforms a random baseline giving random labels to each pixel (see the `Random' line in Table~\ref{tab:scores_all_setups}).

\section{Conclusion}

This paper introduces a novel architecture for semantic change detection in satellite image time series (SITS-SCD) and a detailed analysis of the impact of temporal and spatial domain shifts on this task. Our method outperforms existing baselines for semantic segmentation and binary change segmentation across various evaluation settings, demonstrating its effectiveness in extracting temporal knowledge. Our analysis outlines that spatial domain shift has a significant impact on overall performances, and that temporal domain shift impacts more specifically change detection. However, it also suggests that even in the absence of domain shift, the performance of current methods for SITS-SCD is limited. We believe this is due to the rarity of significant changes, underlining the importance of addressing data scarcity. Overall, our study contributes to advancing SITS-SCD methods and highlights avenues for future research in this area.
\paragraph{Acknowledgments} The work of MA was partly supported by the European Research Council (ERC project DISCOVER, number 101076028). JP was supported in part by the Louis Vuitton/ENS chair on artificial intelligence and the French government under management of Agence Nationale de la Recherche as part of the \textit{Investissements d’avenir} program, reference ANR19-P3IA0001 (PRAIRIE 3IA Institute). This work was granted access to the HPC resources of IDRIS under the allocation 2021-AD011013067 made by GENCI. We thank Antoine Guédon, Loïc Landrieu and Ioannis Siglidis for valuable feedbacks, and Zeynep Sonat Baltaci and Syrine Kalleli for their careful proofreading.
\newpage
{\small
\bibliographystyle{templates/EV24/ieeenat_fullname}
\bibliography{egbib}
}
\clearpage
\setcounter{table}{0}
\renewcommand{\thetable}{A\arabic{table}}
\setcounter{equation}{0}
\renewcommand{\theequation}{A\arabic{equation}}
\setcounter{figure}{0}
\renewcommand{\thefigure}{A\arabic{figure}}

\section*{Appendix A - Dataset details}\label{sec:suppmatdata}

\paragraph{No domain shift setting} As illustrated in Figure~\hyperref[fig:folds]{\ref*{fig:folds}a}, SITS are spatially divided in 4 sub-SITS in this setting. We number all 512$\times$512 quarters of the 1024$\times$1024 SITS from 1 to 4. We follow the 4-fold validation scheme described in Table~\hyperref[tab:folds]{\ref*{tab:folds}a}, where a distinct quarter is kept for test purposes for each fold.

\begin{figure}[t!]
    \centering
    \resizebox{\linewidth}{!}{
    \begin{tabular}{c}
             \includegraphics[width=\linewidth]{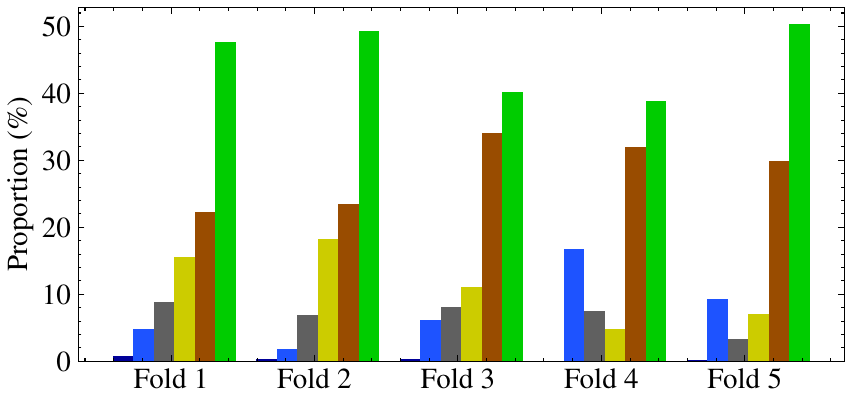}\\
         \includegraphics[width=\linewidth]{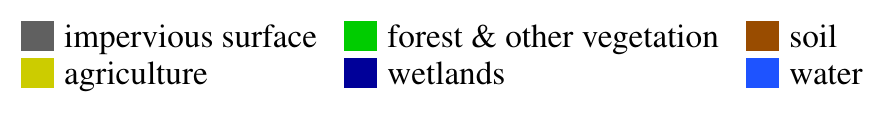} 
    \end{tabular}}
    \caption{Class distribution per fold on DynamicEarthNet in the spatial shift setting.}
    \label{fig:classdistribdnet}
\end{figure}

\begin{figure}[t!]
    \centering
    \resizebox{\linewidth}{!}{
    \begin{tabular}{c}
             \includegraphics[width=\linewidth]{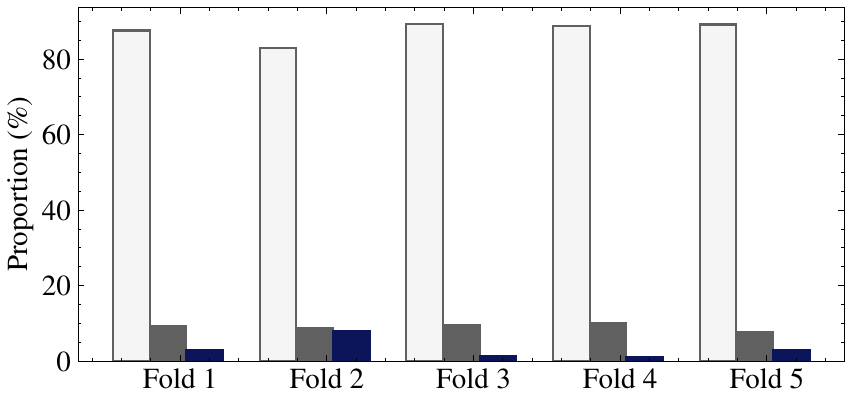}\\
         \includegraphics[width=\linewidth]{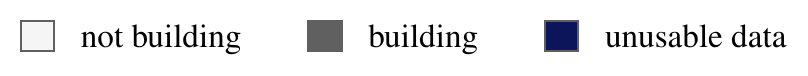} 
    \end{tabular}
    }
    \caption{Class distribution per fold on MUDS in the spatial shift setting.}
    \label{fig:classdistribmuds}
\end{figure}

\paragraph{Spatial shift setting} We detail here the organization of DynamicEarthNet and MUDS's areas of interest into 5 distinct subsets for the spatial shift setting for reproducibility. The no domain shift and temporal shift settings are fully explained in Section~\ref{sec:dom_setup}. In Table~\ref{tab:folds_ids}, we report the compositions of each subset for both datasets. Areas of interest are designated in the table by their unique ID. Note that these subsets have been formed randomly for MUDS (two-class dataset) and in order to have similar class distributions for DynamicEarthNet (multi-class dataset). We show the class distribution for each fold in this setting in Figures~\ref{fig:classdistribdnet} and~\ref{fig:classdistribmuds}. We follow the 5-fold validation scheme described in Table~\hyperref[tab:folds]{\ref*{tab:folds}b} inspired by~\cite{Garnot_2021_ICCV}. 

\begin{table}[h!]
    \centering
    \scriptsize
    \resizebox{\linewidth}{!}{
    \begin{tabular}{ccccc}
    \toprule
    \multicolumn{5}{c}{DynamicEarthNet}\\
    \midrule
    Set 1 & Set 2 & Set 3 & Set 4 & Set 5\\
    \midrule
    2235\_3403\_13 & 2528\_4620\_13 & 1417\_3281\_13 & 1311\_3077\_13 & 1700\_3100\_13\\
    4254\_2915\_13 & 2850\_4139\_13 & 1487\_3335\_13 & 2470\_5030\_13 & 2006\_3280\_13\\
    4421\_3800\_13 & 4240\_3972\_13 & 2415\_3082\_13 & 2832\_4366\_13 & 2029\_3764\_13\\
    4768\_4131\_13 & 4426\_3835\_13 & 2459\_4406\_13 & 4223\_3246\_13 & 2065\_3647\_13\\
    5111\_4560\_13 & 4780\_3377\_13 & 2624\_4314\_13 & 4622\_3159\_13 & 2697\_3715\_13\\
    5989\_3554\_13 & 4856\_4087\_13 & 3002\_4273\_13 & 4806\_3588\_13 & 4791\_3920\_13\\
    6730\_3430\_13 & 5926\_3715\_13 & 3998\_3016\_13 & 5863\_3800\_13 & 4881\_3344\_13\\
    6752\_3115\_13 & 6381\_3681\_13 & 4127\_2991\_13 & 6204\_3495\_13 & 5125\_4049\_13\\
    6810\_3478\_13 & 6813\_3313\_13 & 4169\_3944\_13 & 6466\_3380\_13 & 6468\_3360\_13\\
    6824\_4117\_13 & 7026\_3201\_13 & 4397\_4302\_13 & 7367\_5050\_13 & 6475\_3361\_13\\
    8077\_5007\_13 & 7312\_3008\_13 & 4838\_3506\_13 & 7513\_4968\_13 & 6688\_3456\_13\\
    \midrule
    \multicolumn{5}{c}{MUDS}\\
    \midrule
    Set 1 & Set 2 & Set 3 & Set 4 & Set 5\\
    \midrule
    1446\_2989\_13 & 1549\_3087\_13 & 1736\_3318\_13 & 1327\_3160\_13 & 1429\_3296\_13\\
    1474\_3210\_13 & 2345\_3680\_13 & 2027\_3374\_13 & 1433\_3310\_13 & 1950\_3207\_13\\
    1831\_3648\_13 & 4056\_2688\_13 & 2176\_3279\_13 & 2265\_3451\_13 & 2287\_3888\_13\\
    3041\_4643\_13 & 4102\_2726\_13 & 2383\_3079\_13 & 2528\_4620\_13 & 2309\_3217\_13\\
    4061\_3941\_13 & 4553\_3325\_13 & 2459\_4406\_13 & 3911\_3441\_13 & 2732\_4164\_13\\
    5184\_3399\_13 & 4742\_4450\_13 & 4802\_4803\_13 & 4838\_3737\_13 & 3699\_3757\_13\\
    5342\_3524\_13 & 4815\_3378\_13 & 4816\_3380\_13 & 5753\_3655\_13 & 4196\_2710\_13\\
    6460\_3366\_13 & 4819\_3372\_13 & 5105\_3761\_13 & 5927\_3715\_13 & 4688\_2967\_13\\
    6679\_3549\_13 & 5156\_3514\_13 & 5193\_2903\_13 & 6460\_3370\_13 & 4840\_4088\_13\\
    6813\_3313\_13 & 5916\_3785\_13 & 5759\_3655\_13 & 6468\_3360\_13 & 5557\_3054\_13\\
    6993\_3202\_13 & 6678\_3579\_13 & 6154\_3539\_13 & 6678\_3548\_13 & 6691\_3363\_13\\
    7394\_5018\_13 & 6838\_3742\_13 & 6864\_3345\_13 & 6764\_3347\_13 & 6763\_3346\_13\\
    \bottomrule
    \end{tabular}
    }
    \caption{\textbf{Composition of subsets in the spatial shift setting.} For reproducibility, we share the composition of our subsets in the spatial shift setting. Areas of interest are designated by their unique ID.}
    \label{tab:folds_ids}
\end{table}
\paragraph{Additional details for MUDS} MUDS dataset is not originally designed for semantic segmentation (where annotations are semantic masks) but rather for object detection with polygons labeling all buildings contained in the images of the dataset. To adapt the annotations for our task, we color the interior of polygons in white on a black background to obtain binary semantic mask where 0 corresponds to the class `not building' and 1 to the class `building'. Note that some images of MUDS are not exactly of size 1024$\times$1024 but have a side of size 1023. These images are resized alongside with their semantic mask to the size 1024$\times$1024. Finally, some images from MUDS were acquired on 2017 and 2020: they are discarded for our study. Finally, MUDS comes with `unusable data mask' annotations, \textit{i.e.} binary masks indicating zones where the data cannot realistically be labeled because of cloud, shadows or geo-reference errors. 3.35\% of total pixels are concerned: they are discarded from our metrics. 
\begin{table}[ht!]
    \centering
    \resizebox{0.95\linewidth}{!}{
    \begin{tabular*}{\linewidth}{@{\extracolsep{\fill}}c|ccccc|ccc@{}}
       Fold  & Train & Val & Test & & Fold  & Train & Val & Test \\
       \cmidrule{1-4} \cmidrule{6-9}
        I & 1-2 & 3 & 4 & & I & 1-3 & 4 & 5\\
        II &2-3 & 2 & 3 & & II & 2-4 & 5 & 1\\
        III &3-4 & 1 & 2 & & III & 3-5 & 1 & 2\\
        IV &4-1 & 4 & 1 & & IV & 4-1 & 2 & 3\\
        \multicolumn{4}{c}{} & & V & 5-2 & 3 & 4\\
        \multicolumn{4}{c}{(a) No domain shift setting} & & \multicolumn{4}{c}{(b) Spatial shift setting}\\
        \end{tabular*}
    }
    \caption{$4$- and $5$-fold cross validation schemes for the setting without domain shift and the spatial shift setting. Each line gives the organization of the splits into train, validation and test set for each fold. The temporal domain shift scheme follows the usual train, validation, test single fold procedure.}
    \label{tab:folds}
\end{table}
\setcounter{table}{0}
\renewcommand{\thetable}{B\arabic{table}}
\setcounter{equation}{0}
\renewcommand{\theequation}{B\arabic{equation}}
\setcounter{figure}{0}
\renewcommand{\thefigure}{B\arabic{figure}}

\section*{Appendix B - Implementation details}\label{sec:impl_details}

We train all models on up to 4 NVIDIA GeForce RTX 2080 Ti or NVIDIA V100 GPUs in a data parallel fashion.

\paragraph{UTAE} We used UTAE official Pytorch implementation\footnote{https://github.com/VSainteuf/utae-paps} with default parameters, except for the spatial feature size $D$ that we change according to our different experiments. For positional encoding, we use the number of days after 01/01/2018 at the time of acquisition of the image. For example, an input time series for UTAE weekly corresponding to September 2018 will have [243, 247, 252, 257, 262, 267] as positional encoding vector, since 01/09/2018 is the 244th day since 01/01/2018, and the 5th, 10th, 15th, 20th and 25th days of each months are additionally selected in the weekly setting, following~\cite{Toker_2022_CVPR}.

\paragraph{TSViT} We used TSViT official Pytorch implementation\footnote{https://github.com/michaeltrs/DeepSatModels} with default parameters except for the feature dimension that we set to 512 for fair comparison with other evaluated methods, in terms of trainable parameters, even though \cite{tarasiou2023vits} shows in their supplementary material that the feature size has only a limited impact on performance. We use an input image size of 64$\times$64, since a size 128$\times$128 is exceeding GPU memory capacity. We use the same positional encoding as we do with UTAE.

\paragraph{A2Net and SCanNet} We use A2Net and SCanNet Pytorch implementation available in the Unified Framework for Change Detection\footnote{https://github.com/guanyuezhen/UFCD}. We train over all possible ordered pairs of images. We investigate several inference pairing strategies in Table~\ref{tab:pairing} where the image pairs are defined as follows:
\begin{itemize}
    \item Consecutive: $(\mathbf{x}_{2t},\mathbf{x}_{2t+1})$ with $t\in\{1,...,12\}$;
    \item 12 months apart: $(\mathbf{x}_t,\mathbf{x}_{t+12})$  with $t\in\{1,...,12\}$;
    \item 6 months apart: $(\mathbf{x}_t,\mathbf{x}_{t+6})$ with $t\in\{1,...,6\}\cup\{13, ...18\}$;
    \item Random: $(\mathbf{x}_{t_1},\mathbf{x}_{t_2})$ with $[t_1,t_2]\in\{[0, 15], [1, 13], [2, 9],$ $[3, 23], [4, 10],$ $[5, 14],$ $[6, 18], [7, 19], [8, 21], [11, 12],$ $[16, 17], [20, 22]\}$.
\end{itemize}
It is not clear what is the best strategy to divide a SITS into pairs of image for bi-temporal SITS-SCD approaches since the conclusion drawn depends on what is the considered score and is not always consistent across methods. For these reasons, we stick to the `consecutive' strategy that we deemed is the most natural. 

\paragraph{TSSCD} We use TSSCD official Pytorch implementation\footnote{https://github.com/CUG-BEODL/TSSCD}. Since the method requires pixel time series as input, we flatten 128$\times$128 image patches. At inference all 16,384 resulting pixel time series are used in a batch. During training, we randomly sample 64 pixel time series out of them.

\begin{table}[t!]
  \renewcommand{\arraystretch}{0.9}
  \setlength\tabcolsep{3pt}
  \centering
  \resizebox{\linewidth}{!}{
  \begin{tabular}{lcccccccc}
  \toprule
    \multirow{2}{*}{Pairing} & \multicolumn{4}{c}{A2Net} & \multicolumn{4}{c}{SCanNet}\\
    & \multirow{1}{*}{SCS$\uparrow$} & \multirow{1}{*}{SC$\uparrow$} & \multirow{1}{*}{BC$\uparrow$} & \multirow{1}{*}{mIoU$\uparrow$}& \multirow{1}{*}{SCS$\uparrow$} & \multirow{1}{*}{SC$\uparrow$}& \multirow{1}{*}{BC$\uparrow$} & \multirow{1}{*}{mIoU$\uparrow$}\\
    \midrule
    Consecutive     & \textbf{22.2} & 32.9 & \textbf{11.5} & 47.2 & 24.8 & 35.8 & \textbf{13.9} & 53.0 \\
    12 months apart & 22.0 & 32.9 & 11.2 & 48.2 & \textbf{24.9} & \textbf{36.6} & 13.2 & 54.6\\
    6 months apart  & 21.0 & 31.4 & 10.5 & 48.0 & 23.0 & 33.9 & 12.1 & 53.5\\
    Random          & 21.9 & \textbf{33.0} & 10.8 & \textbf{48.4} & 24.5 & 36.4 & 12.6 & \textbf{54.7} \\
    \bottomrule
  \end{tabular}
  }
  \caption{\textbf{Inference pairing for bi-temporal SITS-SCD.} We report the SCS, SC, BC and mIoU of A2Net and SCanNet for various manners of pairing time stamps at inference.}
  \label{tab:pairing}
\end{table}

\setcounter{table}{0}
\renewcommand{\thetable}{C\arabic{table}}
\setcounter{equation}{0}
\renewcommand{\theequation}{C\arabic{equation}}
\setcounter{figure}{0}
\renewcommand{\thefigure}{C\arabic{figure}}

\section*{Appendix C - Impact of $d$ on performance}\label{sec:impactofd}

\begin{table}[t!]
  \renewcommand{\arraystretch}{0.9}
  \setlength\tabcolsep{3pt}
  \centering
  \resizebox{0.55\linewidth}{!}{
  \begin{tabular}{lcccc}
  \toprule
    & \multirow{1}{*}{SCS$\uparrow$} & \multirow{1}{*}{SC$\uparrow$} & \multirow{1}{*}{BC$\uparrow$} & \multirow{1}{*}{mIoU$\uparrow$}\\
    \midrule
    $d=4$     & \textbf{31.7} & \textbf{41.0} & \textbf{22.4} & \textbf{60.5}\\
    $d=16$   & 31.5 & 40.8 & 22.2 & 60.4\\
    \bottomrule
  \end{tabular}
  }
  \caption{\textbf{Impact of $d$ on performance.} We report the SCS, SC, BC and mIoU of our method in the setting without domain shift on DynamicEarthNet for two values of $d$.}
  \label{tab:impactd}
\end{table}
In the paper, we investigate the impact of the spatial feature size $D$. While the feature maps contain spatial representations, the keys and queries of the attention mechanism store temporal information. Thus, we also analyze the impact of the dimension of keys and queries $d$. We train and evaluate our model on DynamicEarthNet in the setting without domain shift with $d=16$ instead of $d=4$ and report the results in Table~\ref{tab:impactd}. We see little difference if not a very slight decrease for all metrics.

\setcounter{table}{0}
\renewcommand{\thetable}{D\arabic{table}}
\setcounter{equation}{0}
\renewcommand{\theequation}{D\arabic{equation}}
\setcounter{figure}{0}
\renewcommand{\thefigure}{D\arabic{figure}}

\section*{Appendix D - Additional quantitative results}\label{sec:suppmatquant}

\begin{table*}[ht!]
    \definecolor{imp}{RGB}{96, 96, 96}
    \definecolor{not}{RGB}{245, 245, 245}
    \definecolor{agr}{RGB}{204, 204,0}
    \definecolor{for}{RGB}{0, 204,0}
    \definecolor{wet}{RGB}{0, 0, 153}
    \definecolor{soi}{RGB}{153, 76, 0}
    \definecolor{wat}{RGB}{30, 83, 255}
  \makeatletter
  \global\let\oriCT@@do@color\CT@@do@color
  \renewcommand{\arraystretch}{0.95}
  \centering  
  \resizebox{\linewidth}{!}{
  \begin{tabular}{ll|l|l|ccccccc|ccc}
  \toprule
    & \multirow{2}{*}{Method} & \multicolumn{1}{c|}{\multirow{2}{*}{Input type}} & \multicolumn{1}{c|}{\multirow{2}{*}{Strategy}} & \multicolumn{7}{c|}{DynamicEarthNet} & \multicolumn{3}{c}{MUDS}\\
    & & & & \cellcolor{imp!55} imp. surf. & \cellcolor{agr!55}agr. & \cellcolor{for!55}forest & \cellcolor{wet!55}wetlands & \cellcolor{soi!55}soil & \cellcolor{wat!55}water & mean & \cellcolor{not!100}not build. & \cellcolor{imp!55}build. & mean\\
    \midrule
    \parbox[t]{1.3mm}{\multirow{8}{*}{\rotatebox[origin=c]{90}{No domain shift}}} & TSViT monthly & Single image & Single & 26.7 & 46.5 & 72.6 & 13.3 & 56.7 & 87.1 & 50.5 & 91.8 & 28.6 & 60.2\\
    & UTAE monthly & Single image & Single & 33.7 & 53.7 & 75.9 & 11.8 & 59.6 & 87.7 & 53.7 & 92.7 & 41.5 & 67.1 \\
    & TSViT weekly & SITS & Single & 29.5 & 49.4 & 73.9 & 9.4 & 56.8 & 86.6 & 50.9 & --- & --- & ---\\
    & UTAE weekly & SITS & Single & 33.9 & 56.4 & 76.6 & 10.6 & 60.8 & 88.2 & 54.4 & --- & --- & --- \\
    & A2Net & Image pair & Bi & 26.2 & 38.2 & 71.1 & 7.6 & 54.2 & 86.1 & 47.2 & 91.5 & 31.5 & 61.5  \\
    & SCanNet & Image pair & Bi & 30.7 & 53.7 & 74.6 & 12.8 & 58.6 & 87.9 & 53.0 & 92.1 & 37.6 & 64.9 \\
    & TSSCD & Pixel-wise SITS & Multi & 0.2 & 11.5 & 65.6 & 0 & 44.9 & 81.3 & 33.9 & 80.5 & 14.9 & 47.7\\
    \rowcolor{blue!10} \global\let\CT@@do@color & \global\let\CT@@do@color\oriCT@@do@color Ours & SITS & Multi & \textbf{41.6} & \textbf{67.3} & \textbf{80.3} & \textbf{17.9} & \textbf{65.5} & \textbf{90.5} & \textbf{60.5} & \textbf{93.8} & \textbf{50.2} & \textbf{72.0} \\
    \midrule
    \parbox[t]{1.3mm}{\multirow{8}{*}{\rotatebox[origin=c]{90}{Temporal domain shift}}} & TSViT monthly & Single image & Single & 24.4 & 34.5 & 71.1 & 15.7 & 52.5 & 86.1 & 47.3 & 91.4 & 22.2 & 56.8 \\
    & UTAE monthly & Single image & Single & 37.0 & 50.5 & 74.9 & 17.4 & 55.5 & 86.8 & 53.7 & 92.9 & 39.0 & 66.0 \\
    & TSViT weekly & SITS & Single & 28.4 & 42.4 & 72.1 & 22.4 & 54.2 & 88.0 & 51.4 & --- & --- & --- \\
    & UTAE weekly & SITS & Single & 37.4 & 52.8 & 75.1 & 15.8 & 57.9 & 89.0 & 54.7 & --- & --- & --- \\
    & A2Net & Image pair & Bi & 32.8 & 33.2 & 69.5 & 7.1 & 50.6 & 86.8 & 46.7 & 91.5 & 20.8 & 56.1 \\
    & SCanNet & Image pair & Bi & 31.3 & 55.9 & 76.0 & 23.3 & 58.3 & 88.7 & 55.6 & 92.3 & 33.4 & 62.8 \\
    & TSSCD & Pixel-wise SITS & Multi & 0 & 0 & 62.2 & 0 & 39.0 & 75.2 & 29.4 & 83.3 & 15.9 & 49.6 \\
    \rowcolor{blue!10} \global\let\CT@@do@color & \global\let\CT@@do@color\oriCT@@do@color Ours & SITS & Multi & \textbf{42.5} & \textbf{66.3} & \textbf{79.6} & \textbf{30.0} & \textbf{61.8} & \textbf{90.2} & \textbf{61.7} & \textbf{94.0} & \textbf{48.2} & \textbf{71.1} \\
    \midrule
    \parbox[t]{1.3mm}{\multirow{8}{*}{\rotatebox[origin=c]{90}{Spatial domain shift}}} & TSViT monthly & Single image & Single & 12.2 & 9.6 & 55.8 & 0 & 40.4 & 69.4 & 31.2 & 90.8 & 8.9 & 49.8 \\
    & UTAE monthly & Single image & Single & 27.1 & 11.1 & \textbf{64.2} & 0.1 & 43.0 & 76.1 & 36.9 & 92.1 & 33.9 & 63.0 \\
    & TSViT weekly & SITS & Single & 15.1 & 13.4 & 55.7 & \textbf{0.7} & 40.7 & 67.6 & 32.2 & --- & --- & --- \\
    & UTAE weekly & SITS & Single & 29.1 & 15.1 & 63.1 & 0.1 & 43.4 & 76.3 & 37.8 & --- & --- & --- \\
    & A2Net & Image pair & Bi & 24.6 & \textbf{19.9} & 63.4 & 0.1 & 41.7 & \textbf{77.7} & 37.9 & 90.9 & 15.1 & 53.0 \\
    & SCanNet & Image pair & Bi & 22.0 & 14.9 & 64.1 & 0.1 & \textbf{46.5} & 75.9 & 37.3 & 90.1 & 27.6 & 58.8 \\
    & TSSCD & Pixel-wise SITS & Multi & 0.1 & 1.7 & 61.7 & 0 & 39.2 & 34.9 & 22.9 & 78.4 & 8.7 & 43.6 \\
    \rowcolor{blue!10} \global\let\CT@@do@color & \global\let\CT@@do@color\oriCT@@do@color Ours & SITS & Multi & \textbf{30.2} & 16.8 & 62.5 & 0.3 & 45.2 & 76.2 & \textbf{38.5} & \textbf{92.5} & \textbf{39.9} & \textbf{66.2} \\
    \bottomrule
  \end{tabular}
  }
  \caption{\textbf{Per-class IoU.} We report for our method and competing methods the per-class mean intersection-over-union (IoU) in the setting without domain shift on DynamicEarthNet~\cite{Toker_2022_CVPR} and MUDS~\cite{van2021multi}.}
  \label{tab:socres_per_class}
\end{table*}
We report in Table~\ref{tab:socres_per_class} the per-class mean intersection-over-union (IoU) in the setting without domain shift on DynamicEarthNet and MUDS. Our method not only achieves best performance in terms of mean IoU but also on each class in all seetings on MUDS and in the setting without domain shift and in the temporal shift setting on DynamicEarthNet. Note that performance are consistent across all methods, with the hardest class being also the less represented in the datasets (`wetlands' on DynamicEarthNet and `building' on MUDS). An additional comment to be made is that the pixel-wise method TSSCD~\cite{he2024time} performs much worse than all others, once again showing that spatial context-related information leveraged by whole-image methods is crucial for segmentation tasks with SITS. This is a well-known fact in remote sensing, already discussed in~\cite{vincent2023pixel} for example.

\setcounter{table}{0}
\renewcommand{\thetable}{E\arabic{table}}
\setcounter{equation}{0}
\renewcommand{\theequation}{E\arabic{equation}}
\setcounter{figure}{0}
\renewcommand{\thefigure}{E\arabic{figure}}

\section*{Appendix E - Additional qualitative results}\label{sec:suppmatqual}

\begin{figure*}[ht!]
\centering
\setlength{\tabcolsep}{1pt}
\resizebox{\linewidth}{!}{
    \begin{tabular}{cccccccccccc}
        \satimgtssuppmat{input}{14}{}
        \satimgtssuppmat{gt}{14}{}
        \satimgtssuppmat{pred_ours_default}{14}{}
        \\
        \satimgtssuppmat{input}{28}{}
        \satimgtssuppmat{gt}{28}{}
        \satimgtssuppmat{pred_ours_default}{28}{}
        \\
        \satimgtssuppmat{input}{39}{}
        \satimgtssuppmat{gt}{39}{}
        \satimgtssuppmat{pred_ours_default}{39}{}
        \\
        \satimgtssuppmat{input}{40}{}
        \satimgtssuppmat{gt}{40}{}
        \satimgtssuppmat{pred_ours_default}{40}{}
        \multicolumn{12}{c}{\includegraphics[width=\linewidth]{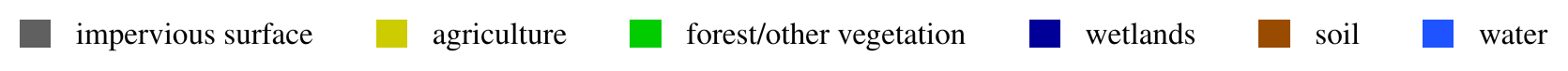}}
    \end{tabular}
    }
    \caption{{\bf Qualitative segmentation results on DynamicEarthNet.} We show the segmentation maps predicted by our model in the setting without domain shift for four randomly selected input SITS from DynamicEarthNet. For each SITS, we show the monthly input time series from January to December 2019 (top row), the corresponding ground truth (middle row) and the predictions (bottom row).}
    \label{fig:suppmat_denet_ts}
\end{figure*}

\begin{figure*}[ht!]
\centering
\setlength{\tabcolsep}{1pt}
\resizebox{\linewidth}{!}{
    \begin{tabular}{cccccccccccc}
        \satimgtssuppmat{input}{8}{muds_}
        \satimgtssuppmat{gt}{8}{muds_}
        \satimgtssuppmat{pred}{8}{muds_ours_default_}
        \\
        \satimgtssuppmat{input}{36}{muds_}
        \satimgtssuppmat{gt}{36}{muds_}
        \satimgtssuppmat{pred}{36}{muds_ours_default_}
        \\
        \satimgtssuppmat{input}{55}{muds_}
        \satimgtssuppmat{gt}{55}{muds_}
        \satimgtssuppmat{pred}{55}{muds_ours_default_}
        \\
        \satimgtssuppmat{input}{58}{muds_}
        \satimgtssuppmat{gt}{58}{muds_}
        \satimgtssuppmat{pred}{58}{muds_ours_default_}
        \multicolumn{12}{c}{\includegraphics[width=0.5\linewidth]{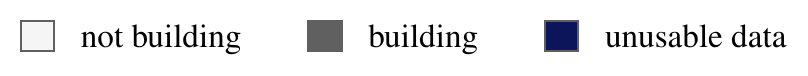}}
    \end{tabular}
    }
    \caption{{\bf Qualitative segmentation results on MUDS.} We show the segmentation maps predicted by our model in the setting without domain shift for four randomly selected input SITS from MUDS. For each SITS, we show the monthly input time series from January to December 2019 (top row), the corresponding ground truth (middle row) and the predictions (bottom row).}
    \label{fig:suppmat_muds_ts}
\end{figure*}

We show in Figures~\ref{fig:suppmat_denet_ts} and~\ref{fig:suppmat_muds_ts} the semantic predictions of our model for 12-month time series on DynamicEarthNet and MUDS respectively in the setting without domain shift. Our architecture is able to capture changes while jointly processing the whole time series at once. Note how our model can classify accurately buildings masked as `unusable data' in MUDS dataset.

\clearpage

\clearpage

\end{document}